\documentclass{article}
\usepackage{spconf,amsmath,graphicx,hyperref}

\usepackage{graphicx}
\usepackage{amsmath}
\usepackage{amssymb}
\usepackage{multirow}
\usepackage[table,xcdraw]{xcolor}
\usepackage{booktabs}
\usepackage{amssymb}
\usepackage{pifont}
\usepackage{orcidlink}
\hypersetup{hidelinks}

\title{Learning Reference-Guided Exposure Correction with Hybrid Illumination Characteristics}
\name{{Hao Ren~\orcidlink{0009-0000-6755-8971}, 
	  Zetong Bi~\orcidlink{0009-0007-3435-1940},
	Zhaoliang Wan~\orcidlink{0000-0003-0675-0665}, 
	Hui Cheng~\orcidlink{0000-0003-2579-7004}$^\star$}\thanks{$^\star$Corresponding to chengh9@mail.sysu.edu.cn. This work was supported by the National Natural Science Foundation of China (U22A2095).}
    }

\address{School of Computer Science and Engineering, Sun Yat-sen University, Guangzhou, China}
\begin{document}
\ninept
%


\maketitle              

\begin{abstract}
We present HICNet, a reference-guided exposure correction framework. A lightweight, content-agnostic encoder distils each image into a compact illumination embedding capturing regional brightness, edge contrast, and higher-order luminance moments. The embedding difference between a source and its reference drives a multi-scale modulation network that combines FiLM-based global adjustment with Photometric Channel Rebalancing for fine-grained, illumination-aware spectral gating, producing exposure-matched outputs while faithfully preserving scene details. A cross-batch contrastive loss orders the illumination manifold, bolstering robustness to diverse lighting conditions. Trained without ground truth or intrinsic decomposition, HICNet attains better accuracy on public benchmarks and generalizes well to entirely unseen scenes.

\end{abstract}


\section{Introduction}

Exposure correction is a fundamental task in computational photography and low-level vision, aiming to restore under- or over-exposed images to visually natural appearances. In real-world scenarios, factors such as uneven lighting, suboptimal camera settings, or limited dynamic range often degrade brightness, leading to detail loss, low contrast, and perceptual discomfort~\cite{afifi2021learning, eyiokur2022exposure, ren2025layer}. Robust exposure correction is therefore essential for improving image quality.

Despite its significance, exposure correction remains a challenge. Existing methods include two main categories: supervised learning with paired expert ground truth~\cite{chen2018learning, afifi2021learning, ignatov2017dslr, wang2019underexposed} and decomposition-based enhancement grounded in Retinex theory~\cite{wei2018deep}. Supervised approaches learn direct mappings from under- or over-exposed inputs to properly exposed references, but heavily depend on carefully annotated datasets. Such datasets are costly to obtain and may not generalize well across varied illumination conditions. 
On the other hand, decomposition-based methods model images by disentangling reflectance and illumination but typically rely on strong priors, such as spatial smoothness or strict component separation, which limit their adaptability and robustness in complex scenes with varied content and illumination.

To overcome these limitations, recent studies have explored unsupervised exposure enhancement based on distribution alignment or auxiliary exposure metrics~\cite{guo2020zero, wang2022local}. However, many of these approaches still rely on hand-crafted priors, require ground truth images, or implicitly assume specific illumination structures. 
A critical, often overlooked challenge in unsupervised settings is the domain gap between the source and reference images. Merely aligning global statistics can be detrimental if the semantic content differs significantly. Therefore, the key to robust reference-guided correction lies not just in alignment, but in effectively decoupling illumination style from semantic content without relying on expert supervision.

In this paper, we propose HICNet, a reference-guided framework centered on the Content-Agnostic Exposure Encoder (CAEE).
Motivated by the need to disentangle lighting from content, the CAEE is explicitly designed to capture exposure style in terms of regional brightness, gradients, and high-order moments, while ignoring semantic details. This mechanism allows the network to extract the desired illumination from a reference and transfer it to the source without being misled by content discrepancies. Specifically, global exposure is adjusted via feature-wise linear modulation (FiLM), while fine-grained, spectrum-aware gains are produced by Photometric Channel Rebalancing (PCR), which converts the illumination difference into per-channel gating to emphasise exposure-sensitive feature bands. To further structure the representation, a cross-batch contrastive loss orders the illumination manifold, thereby improving robustness to diverse lighting conditions. Extensive experiments on standard benchmarks show that HICNet delivers better or competitive performance and generalises well across scenes and illumination conditions, all without relying on expert supervision.

\section{Related Work}
\label{sec:related}

\textbf{Exposure Correction}. 
Classic exposure correction relied on global or local intensity remapping such as histogram equalization, gamma adjustment, and CLAHE~\cite{celik2011contextual, park2008contrast, pizer1987adaptive}. These handcrafted schemes lack scene awareness and require careful parameter adjustment for images. 
Retinex-inspired decompositions factor an image into illumination and reflectance~\cite{wei2018deep}, but this ill-posed separation depends on strong, scene-specific priors.  
Deep networks trained on paired low/high-exposure data achieve impressive fidelity~\cite{chen2018learning, afifi2021learning, parab2023image}, yet require costly ground-truth collection and generalize poorly outside the training domain.  
Unsupervised alternatives (e.g., Zero-DCE, QuadPrior, UEC)~\cite{guo2020zero, wang2024zero, cui2024unsupervised} remove the pairing constraint by leveraging adversarial learning or analytical curves, but usually focus on a narrow exposure range or embed rigid assumptions, lacking an explicit, controllable representation of exposure style.

\noindent
\textbf{Style-Based Enhancement and Feature Modulation}. Global style embedding injected via FiLM or AdaIN layers steers feature statistics for high-level synthesis~\cite{perez2018film, huang2017arbitrary, ren2025prior}. Contrastive objectives further shape unconditional embeddings~\cite{chen2020simple} and align cross-domain representations, as popularized by CLIP~\cite{radford2021learning}.  
In low-level vision, however, these ideas are seldom exploited to match exposure between unpaired images. Existing unsupervised exposure methods provide limited control and no principled mechanism for cross-image brightness alignment. We fill this gap by coupling a histogram-based illumination encoder with FiLM modulation, enabling fine-grained exposure transfer without paired ground truth.

\section{Method}
\label{sec:method}
\begin{figure}[t]
    \centering
    \includegraphics[width=\linewidth]{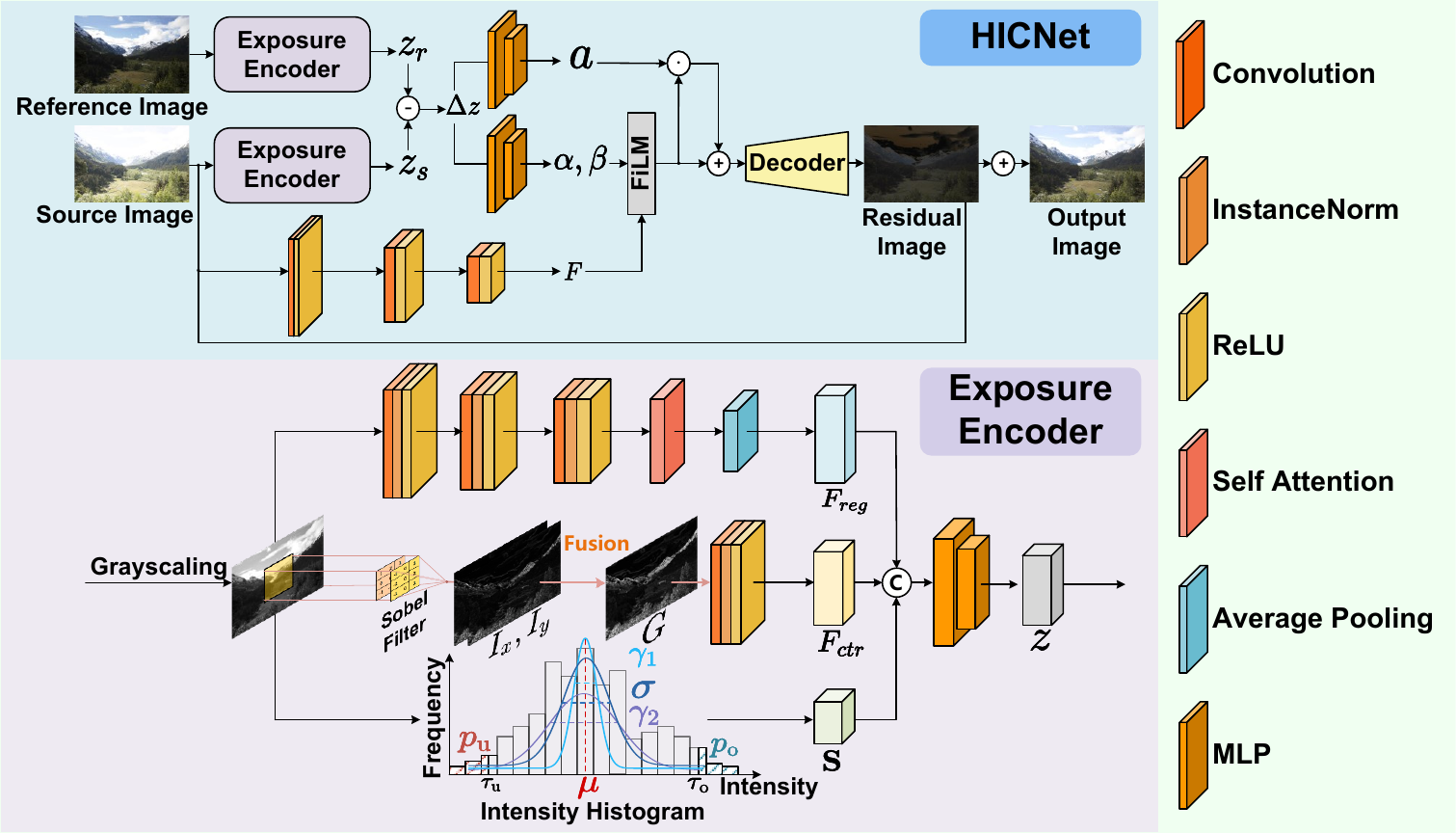}
    \vspace{-15pt}
    \caption{Overview of HICNet. Given a source image $I_s$ and a reference image $I_r$, each image is mapped by a content-agnostic encoder to a compact illumination descriptor. Their difference summarizes the exposure shift and conditions a U-shaped backbone via feature-wise modulation and photometric channel rebalancing to predict a residual, which is added to $I_s$ to produce the corrected result. The bottom panel outlines the encoder’s three components, regional brightness, gradient-based contrast, and global histogram statistics, while the examples on the right illustrate how the reference guides exposure and contrast without altering scene structure.}

    \vspace{-10pt}
    \label{fig:pipline}
\end{figure}

Figure~\ref{fig:pipline} gives an overview of the proposed pipeline.  
The system comprises two learnable modules.  
First, a \emph{Content-Agnostic Exposure Encoder} compresses any input image into a low-dimensional code that summarizes its exposure style while remaining oblivious to scene semantics.  
Second, a \emph{Multi-Scale Modulation Network} injects the difference between the embedding of a source image \(I_s\) and a reference image \(I_r\) into a U-shaped backbone, thereby producing an exposure-corrected output \(\hat I\).  
All components are trained jointly and end-to-end without paired supervision or explicit reflectance–illumination decomposition.

\subsection{Problem Formulation}

Let \(I_s,I_r\in\mathbb{R}^{3\times H\times W}\) denote the source image with undesirable exposure and a reference image exhibiting the desired brightness distribution, respectively; the two images may depict entirely different scenes.  
Our goal is to learn a mapping:
\begin{equation}
    f:(I_s,I_r)\longrightarrow \hat I,
\end{equation}
such that the prediction \(\hat I\) retains the content of \(I_s\) while aligning its global exposure statistics with those of \(I_r\).  
During training, no ground-truth triplets \((I_s,I_r,\hat I^{\!*})\) are available; the network must therefore rely solely on information present in the two inputs.

\subsection{Content-Agnostic Exposure Encoder}
\label{subsec:encoder}

A key challenge is to describe exposure in a way that is both \emph{informative}, capturing where and how light is distributed, and \emph{generic}, robust across scenes and free of object semantics.  
To this end, we combine three complementary signals.

\textbf{(1) Region-pooled brightness.}
The RGB image is first converted to grayscale,  
\(I_{\text{gray}} = 0.2989R+0.5870G+0.1140B\).  
Three shallow Conv–InstanceNorm–ReLU blocks followed by a single self-attention layer gather long-range brightness correlations without learning object shapes.  
The resulting feature map is down-sampled with a \(2\times2\) adaptive average pool, yielding a 256-D vector \(F_{\text{reg}}\) that indicates where bright or dark areas are roughly located.

\textbf{(2) Gradient-based local contrast.}
Over-exposed regions often erase edges, whereas under-exposed shadows retain noise-amplified gradients.  
Fixed Sobel kernels compute horizontal and vertical derivatives \((I_x,I_y)\); their magnitude
\(
G=\sqrt{I_x^{2}+I_y^{2}}
\)
is subsequently compressed by a tiny CNN into a 32-D vector \(F_{\text{ctr}}\).  
Because the Sobel operator is parameter-free, the representation remains content-neutral yet sensitive to exposure-induced contrast changes.

\textbf{(3) Global histogram moments.}
Let \(N=H\!\times\!W\) be the number of pixels.  
We measure the mean \(\mu\), standard deviation \(\sigma\), skewness \(\gamma_1\), and kurtosis \(\gamma_2\) of \(I_{\text{gray}}\), as well as the percentages of severely under-exposed and over-exposed pixels:
\begin{equation}
p_{\mathrm{u}}=\frac{1}{N}\sum\nolimits_{x}\mathbf 1[I_{\text{gray}}(x)<\tau_{\mathrm{u}}],\quad
p_{\mathrm{o}}=\frac{1}{N}\sum\nolimits_{x}\mathbf 1[I_{\text{gray}}(x)>\tau_{\mathrm{o}}],
\end{equation}

with thresholds \(\tau_{\mathrm{u}}{=}0.05\) and \(\tau_{\mathrm{o}}{=}0.95\) in the normalised range \([0,1]\).  
Stacking these six values forms the statistic vector \(\mathbf S\in\mathbb{R}^{6}\), which concisely characterises the histogram’s shape and saturation extremes.

\textbf{Descriptor formation.}
Concatenating the three branches and projecting through a two-layer multilayer perceptron \(\Phi\) yields the illumination descriptor:
\begin{equation}
    z=\Phi\!\Bigl(\bigl[\,F_{\text{reg}};\,F_{\text{ctr}};\,\mathbf S\,\bigr]\Bigr)\in\mathbb{R}^{66}.
\end{equation}
Because every component is either shallow or statistic-based, the final code remains lightweight (only 66 numbers) and largely free of semantic entanglement.

\subsection{Multi-Scale Exposure Modulation}
\label{subsec:modnet}

The correction network adopts a three-stage U-Net encoder–decoder whose feature maps are denoted \(\{F_1,F_2,F_3\}\) with channel dimensions \(\{32,64,64\}\).
First, we compute the illumination shift: 
\begin{equation}
    \Delta z=z_r-z_s,
    \label{eq:deltaz_repeat}
\end{equation}
where \(z_s\) and \(z_r\) are the descriptors of \(I_s\) and \(I_r\).  
Vector \(\Delta z\) encodes how exposure must change to move from the source to the reference domain.

\textbf{Feature-wise linear modulation.}
For every scale \(i\in\{1,2,3\}\) a small MLP predicts FiLM parameters \([\alpha_i,\beta_i]=\mathrm{MLP}_i(\Delta z)\in\mathbb{R}^{2C_i}\).  
These parameters perform an affine transformation:
\begin{equation}
    \widetilde F_i=\alpha_i\odot F_i+\beta_i,
    \label{eq:film_repeat}
\end{equation}
where \(\odot\) denotes channel-wise multiplication after broadcasting \(\alpha_i,\beta_i\) to the spatial size of \(F_i\).  The operation is content-preserving because it neither mixes spatial locations nor alters feature topology.

\textbf{Photometric Channel Rebalancing (PCR).}
While Eq.~\eqref{eq:film_repeat} modulates every feature channel, their contributions to perceived luminance differ across scenes. PCR translates the illumination-shift embedding $\Delta z$ and coarse content statistics into per-channel gates:
\begin{equation}
    a_i=\sigma\!\left(W_i\,\Delta z+b_i\right)\in(0,1)^{C_i},
\end{equation}
where $(W^{\!\text{attn}}_{i},b^{\text{attn}}_{i})$ are linear weights and \(\sigma(\cdot)\) is the sigmoid activation.  
The rebalanced feature map is then:
\begin{equation}
    \widehat F_i=(1+a_i)\odot\widetilde F_i,
\end{equation}
so channels assigned higher photometric weights \(a_i\) are proportionally amplified, while less relevant channels remain near their original scale. This PCR block thus converts illumination discrepancy into selective spectral gain, enabling the network to emphasise exposure-sensitive bands and suppress redundancies.

\textbf{Decoder and reconstruction.}
The decoder upsamples \(\{\widehat F_i\}\) in a top-down manner, concatenating with lower-level features via skip connections.  
A final \(3\times3\) convolution produces a residual map \(R\), and the corrected image is obtained as: $\hat I = \operatorname{clip}(I_s+R,\;0,1),$
where \(\operatorname{clip}\) ensures that pixel intensities stay within the valid range.

Injecting \(\Delta z\) at multiple resolutions enables simultaneous alignment of coarse global brightness and delicate local nuances, while the attention gate suppresses unnecessary adjustments in illumination-irrelevant channels.

\subsection{Training Objectives}
\label{subsec:loss}

Learning is driven by three complementary losses:

\textbf{Reconstruction loss.}
A direct \(\ell_1\) difference between the prediction and the reference $\mathcal L_{\text{pix}}=\lVert\hat I-I_r\rVert_1$
encourages the average brightness and colour balance of \(\hat I\) to approach \(I_r\).

\textbf{Dark-channel prior loss.}
The dark channel \(D(I)\) of an RGB image is the minimum intensity inside a local \(16\times16\) neighbourhood across all colour channels.  
Matching dark channels, $\mathcal L_{\text{dc}}=\lVert D(\hat I)-D(I_r)\rVert_1,$
sharpens local contrast and prevents blown-out highlights by drawing on the de-hazing principle~\cite {he2010single}.

\textbf{Contrastive illumination loss.}
We normalize the L2-exposed descriptors \(z\) using a multi-positive NT-Xent loss.
To avoid circular mining, positives are defined in a momentum (EMA) teacher space.
For an anchor \(i\), let \(P(i)=\{a\neq i:\ \|z_i^t-z_a^t\|_2\le\delta_i\}\), where \(\delta_i\) is the \(q\)-th percentile of teacher-space distances within the batch plus a memory queue; anchors with \(P(i)=\varnothing\) are skipped.
The loss is:
\begin{equation}
\mathcal{L}_i\!
=\!-\frac{1}{|P(i)|}\!\sum_{p\in P(i)}\!
\log\frac{\exp\!\big(z_i^{s\top}z_p^t/\tau\big)}
{\sum_{a\neq i}\exp\!\big(z_i^{s\top}z_a^t/\tau\big)},
\mathcal{L}_{\text{ctr}}=\tfrac{1}{|\mathcal{B}|}\sum_i\mathcal{L}_i.
\end{equation}

\textbf{Full objective.}
The network is trained end-to-end using:
\begin{equation}
    \mathcal L=\mathcal L_{\text{pix}} + \lambda_{dc}\,\mathcal L_{\text{dc}} + \lambda_{ctr}\mathcal L_{\text{ctr}},
\end{equation}
where $\lambda_{dc}$ and $\lambda_{ctr}$ denote the respective weights of the loss functions.

\section{Experiments}
\subsection{Experiment Settings}
\textbf{Dataset.} We evaluate our method on two datasets: the MSEC Dataset from~\cite{afifi2021learning}. The MSEC dataset consists of images captured under varying exposure conditions, enabling us to test the robustness of our method across different illumination levels. Additionally, to validate generalization, we test the model trained on the exposure correction dataset on the LOL dataset~\cite{cai2018learning}, which is specifically designed for low-light image enhancement.

\noindent
\textbf{Implementation Details.} 
We train for 100 epochs at a constant learning rate of 0.0002 (Adam optimizer), followed by 100 epochs of linear decay. The batch size is 32. We resize the images to $512\times512$ pixels. We set $\lambda_{dc}=0.5$, $\lambda_{ctr}=0.1$, and \(\tau=0.07\). We chose the first image in the MSEC training set as the reference image.

\noindent
\textbf{Evaluation Metrics.} 
To evaluate the overall quality of the generated images, we compare the output to the ground truth images using standard metrics such as Peak Signal-to-Noise Ratio (PSNR) and Structural Similarity Index (SSIM) [35]. For the generalization study, we also introduce the Natural Image Quality Evaluator (NIQE) and the Perceptual Index (PI), two no-reference image quality evaluation metrics, to assess image quality. 

\begin{figure}[t]
    \centering
    \includegraphics[width=\linewidth]{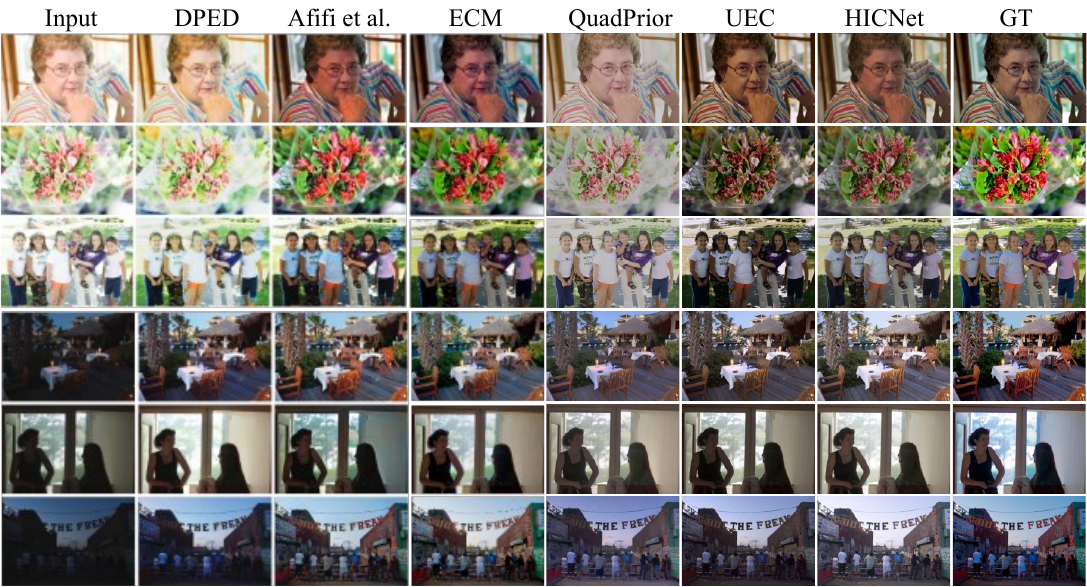}
    \vspace{-15pt}
    \caption{Exposure correction results on MSEC dataset. We take images from~\cite{eyiokur2022exposure} and compare with our model.}
    \label{fig:MSEC_results}
    \vspace{-10pt}
\end{figure}
\begin{table}[t]
\centering
\vspace{-5pt}
\caption{Quantitative comparison on the MSEC~\cite{afifi2021learning} test set. "Sup." indicates whether the method is supervised (S) or unsupervised (U). Training-free methods are marked by $\ast$. Algorithms designed for low-light image enhancement are marked in gray.}
\label{table:results_all}
\renewcommand{\arraystretch}{0.8}
\setlength{\tabcolsep}{1pt}
\scalebox{0.85}{
\begin{tabular}{l c ll|ll|ll}
\toprule
\multirow{2}{*}{\textbf{Method}} & \multirow{2}{*}{\textbf{Sup.}} & \multicolumn{2}{c|}{\textbf{Overexposure}} & \multicolumn{2}{c|}{\textbf{Underexposure}} & \multicolumn{2}{c}{\textbf{Average}} \\
& & \textbf{PSNR↑} & \textbf{SSIM↑} & \textbf{PSNR↑} & \textbf{SSIM↑} & \textbf{PSNR↑} & \textbf{SSIM↑} \\
\midrule

HDR CNN~\cite{eilertsen2017hdr} & S & 16.08 & 0.680 & 18.47 & 0.698 & 17.03 & 0.687 \\
DPED (BB)~\cite{ignatov2017dslr} & S & 16.44 & 0.662 & \cellcolor[HTML]{fff2cc}20.06 & 0.685 & 17.89 & 0.671 \\
DPE (HDR)~\cite{chen2018deep} & S & 15.41 & 0.589 & 17.40 & 0.673 & 16.21 & 0.623 \\
DPE (U-FiveK)~\cite{chen2018deep} & S & 15.45 & 0.640 & 18.50 & 0.677 & 16.67 & 0.655 \\
DPE (S-FiveK)~\cite{chen2018deep} & S & 16.04 & 0.661 & \cellcolor[HTML]{f4cccc}19.72 & 0.702 & 17.51 & 0.677 \\
\rowcolor[HTML]{D5D5D5}RetinexNet~\cite{wei2018deep} & S & 11.06 & 0.600 & 12.49 & 0.619 & 11.63 & 0.607 \\
\rowcolor[HTML]{D5D5D5}Deep UPE~\cite{wang2019underexposed} & S & 11.01 & 0.573 & 19.11 & \cellcolor[HTML]{f4cccc}0.741 & 14.25 & 0.640 \\
Afifi et al. w/o $\mathcal{L}_{\text{adv}}$~\cite{afifi2021learning} & S & \cellcolor[HTML]{fff2cc}19.35 & \cellcolor[HTML]{fff2cc}0.737 & 19.69 & \cellcolor[HTML]{fff2cc}0.742 & \cellcolor[HTML]{fff2cc}19.48 & \cellcolor[HTML]{fff2cc}0.739 \\
Afifi et al. w/ $\mathcal{L}_{\text{adv}}$~\cite{afifi2021learning} & S & \cellcolor[HTML]{f4cccc}19.20 & \cellcolor[HTML]{f4cccc}0.728 & 19.65 & 0.737 & \cellcolor[HTML]{f4cccc}19.38 & \cellcolor[HTML]{f4cccc}0.731 \\
ECM~\cite{eyiokur2022exposure} & S & \cellcolor[HTML]{c6efce}20.99 & \cellcolor[HTML]{c6efce}0.881 & \cellcolor[HTML]{c6efce}20.71 & \cellcolor[HTML]{c6efce}0.873 & \cellcolor[HTML]{c6efce}20.87 & \cellcolor[HTML]{c6efce}0.877 \\
\midrule

HE~\cite{gonzalez2001digital}$\ast$ & U & 16.58 & 0.683 & 16.58 & 0.679 & 16.58 & 0.682 \\
CLAHE~\cite{reza2004realization}$\ast$ & U & 14.81 & 0.589 & 16.35 & 0.621 & 15.43 & 0.602 \\
WVM~\cite{fu2016weighted}$\ast$ & U & 13.50 & 0.657 & \cellcolor[HTML]{fff2cc}18.62 & 0.735 & 15.55 & 0.688 \\
\rowcolor[HTML]{D5D5D5}LIME~\cite{guo2016lime}$\ast$ & U & 10.49 & 0.582 & 14.64 & 0.671 & 12.15 & 0.618 \\
\rowcolor[HTML]{D5D5D5}HQEC~\cite{zhang2018high}$\ast$ & U & 12.88 & 0.638 & 16.91 & 0.706 & 14.49 & 0.666 \\
\rowcolor[HTML]{D5D5D5}SCI (Official)~\cite{ma2022toward} & U & 6.60 & 0.523 & 9.73 & 0.638 & 7.64 & 0.561 \\
\rowcolor[HTML]{D5D5D5}SCI (Retraining)~\cite{ma2022toward} & U & \cellcolor[HTML]{fff2cc}18.38 & \cellcolor[HTML]{f4cccc}0.795 & 16.35 & 0.736 & \cellcolor[HTML]{fff2cc}17.71 & 0.776 \\
\rowcolor[HTML]{D5D5D5}Zero-DCE~\cite{guo2020zero} & U & 11.02 & 0.520 & 14.96 & 0.594 & 12.60 & 0.549 \\
\rowcolor[HTML]{D5D5D5}QuadPrior~\cite{wang2024zero} & U & 16.99 & 0.778 & \cellcolor[HTML]{f4cccc}17.95 & \cellcolor[HTML]{f4cccc}0.79 & 17.38 & \cellcolor[HTML]{f4cccc}0.783 \\
UEC~\cite{cui2024unsupervised} & U & \cellcolor[HTML]{f4cccc}17.64 & \cellcolor[HTML]{fff2cc}0.808 & 17.42 & \cellcolor[HTML]{c6efce}0.812 & \cellcolor[HTML]{f4cccc}17.55 & \cellcolor[HTML]{fff2cc}0.809 \\
\textbf{HICNet (Ours)} & U & \cellcolor[HTML]{c6efce}18.91 & \cellcolor[HTML]{c6efce}0.817 & \cellcolor[HTML]{c6efce}19.04 & \cellcolor[HTML]{fff2cc}0.811 & \cellcolor[HTML]{c6efce}18.96 & \cellcolor[HTML]{c6efce}0.815 \\
\bottomrule
\end{tabular}
}
\vspace{-10pt}
\end{table}

\begin{figure}[t]
    \centering
    \includegraphics[width=\linewidth]{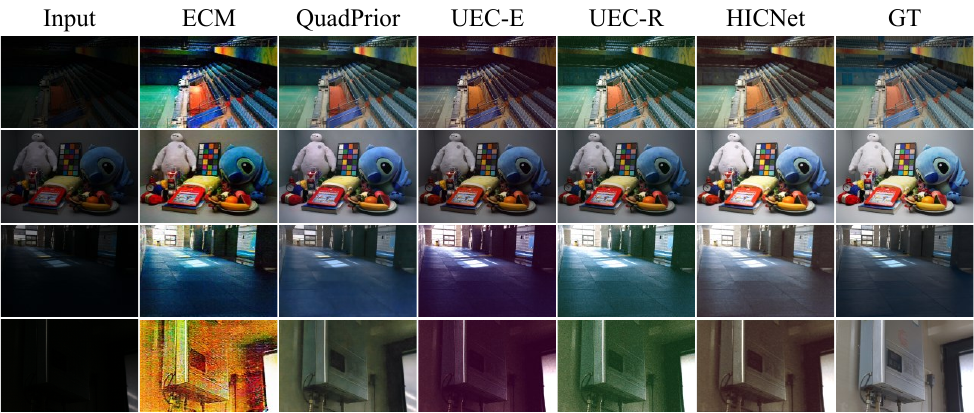}
    \vspace{-15pt}
    \caption{A visualization comparing the generalizability performance of HICNet (our method) with state-of-the-art methods.}
    \label{fig:LOL_results}
    \vspace{-10pt}
\end{figure}
\begin{table}[t]
\centering
\vspace{-5pt}
\caption{
Generalization performance on LOL dataset~\cite{cai2018learning}. 
Methods are pretrained on either the MSEC dataset (denoted by "-E") or the Radiometry Correction dataset ("-R"). 
}
\label{table:results_generalization}
\renewcommand{\arraystretch}{0.8}
\scalebox{0.9}{
\begin{tabular}{lcccc}
\toprule
\textbf{Method} & \textbf{PSNR $\uparrow$} & \textbf{SSIM $\uparrow$} & \textbf{NIQE $\downarrow$} & \textbf{PI $\downarrow$} \\
\midrule
ECM~\cite{eyiokur2022exposure} & 15.3733 & 0.6333 & \textbf{4.4677} & \textbf{3.2860} \\
UEC-E~\cite{cui2024unsupervised} & 14.9568 & 0.5896 & 8.1772 & 4.7456 \\
UEC-R~\cite{cui2024unsupervised} & 17.5020 & 0.6539 & 8.3440 & 4.8813 \\
QuadPrior~\cite{wang2024zero} & 18.6292 & 0.8116 & 5.4426 & 3.9716 \\
\textbf{HICNet (Ours)} & \textbf{19.3290} & \textbf{0.8174} & 4.7537 & 4.4938 \\
\bottomrule
\end{tabular}
}
\vspace{-15pt}
\end{table}

\begin{figure}[t]
    \centering
    \includegraphics[width=\linewidth]{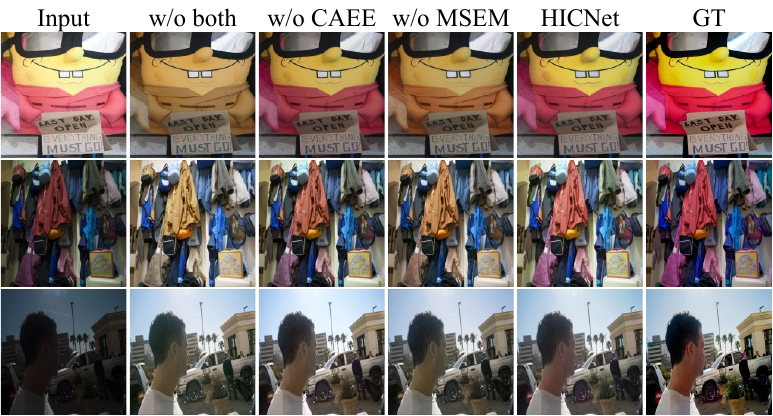}
    \vspace{-15pt}
    \caption{Visualization of ablation study on MSEC~\cite{afifi2021learning} test set.}
    \label{fig:ablation}
    \vspace{-10pt}
\end{figure}
\begin{table}[t]
\centering
\renewcommand{\arraystretch}{0.8}
\vspace{-5pt}
\caption{Ablation studies on MSEC~\cite{afifi2021learning} test set.}
\scalebox{0.9}{
\begin{tabular}{c c c c}
\toprule
\textbf{CAEE} & \textbf{MSEM} & \textbf{PSNR$\uparrow$} & \textbf{SSIM$\uparrow$} \\
\midrule
\ding{51} & \ding{51} & \textbf{18.962} & \textbf{0.815} \\
\ding{55} & \ding{51} & 18.503 & 0.792 \\
\ding{51} & \ding{55} & 18.122 & 0.770 \\
\ding{55} & \ding{55} & 17.935 & 0.755 \\
\bottomrule
\end{tabular}
}
\vspace{-10pt}
\label{tab:ablation}
\end{table}

\subsection{Results on MSEC Dataset}

\textbf{Qualitative Evaluation}. Figure~\ref{fig:MSEC_results} presents qualitative comparisons on the MSEC dataset~\cite{afifi2021learning}. HICNet effectively corrects exposure while preserving natural colors and fine details. Compared to leading unsupervised baselines like QuadPrior~\cite{wang2024zero} and UEC~\cite{cui2024unsupervised}, our method achieves superior visual alignment with the ground truth, particularly in terms of color fidelity and exposure balance. While supervised methods such as ECM~\cite{eyiokur2022exposure} may offer slight advantages in specific cases, HICNet delivers highly competitive visual quality without requiring paired supervision, validating the efficacy of our content-agnostic design.

\noindent
\textbf{Quantitative Evaluation}. Table~\ref{table:results_all} summarizes the quantitative performance on the MSEC test set~\cite{afifi2021learning} in terms of PSNR and SSIM. The best, second-best, and third-best results are highlighted in green, yellow, and red, respectively. HICNet consistently outperforms existing unsupervised methods and remains competitive with supervised approaches across diverse conditions.

In overexposure scenarios, HICNet surpasses unsupervised baselines (e.g., Zero-DCE~\cite{guo2020zero}, HE~\cite{gonzalez2001digital}) and achieves parity with supervised models like DPED~\cite{ignatov2017dslr}, demonstrating robustness in high-brightness recovery. Similarly, in underexposure scenarios, our method outperforms widely used techniques such as SCI~\cite{ma2022toward} and QuadPrior~\cite{wang2024zero}, while closely matching supervised DPE~\cite{chen2018deep} variants. Overall, these results confirm that our unsupervised, multi-scale modulation architecture provides a robust, balanced solution for real-world applications where paired training data is unavailable.

\subsection{Generalization Performance}

Figure~\ref{fig:LOL_results} provides a visual comparison of exposure‐correction results produced by other comparison methods when models trained on the MSEC dataset~\cite{afifi2021learning} are evaluated on the unseen low-light LOL dataset.  
Across all scenes, HICNet delivers corrections that are perceptually closest to the ground truth: colours remain faithful (rows~1–2), shadow recovery is free of colour shifts (row~3), and structural details are preserved without texture over-smoothing (row~4).  

By contrast, the supervised baseline ECM~\cite{eyiokur2022exposure} occasionally amplifies noise and introduces chromatic artefacts in heavily under-exposed regions (row~4), while QuadPrior~\cite{wang2024zero} tends to over-saturate bright areas and suffer from hue distortions (rows~1 and~2). Unsupervised competitors UEC-E and UEC-R often leave residual colour casts or haze-like veiling, indicating limited domain transfer.  

These qualitative observations corroborate the quantitative gains reported in Table~\ref{table:results_generalization}: the semantic-agnostic illumination descriptor and multi-scale modulation network of HICNet effectively adapt to a markedly different low-light distribution, yielding visually compelling results without any additional fine-tuning.

\subsection{Ablation Study}
We ablate the two key components of HICNet: the content-agnostic exposure encoder (CAEE), which compresses regional brightness, contrast, and histogram statistics into a lightweight illumination code, and the multi-scale exposure modulation (MSEM), which injects the source–reference illumination shift via FiLM and photometric channel rebalancing across the U-shape network.    

Table~\ref{tab:ablation} shows that the complete model (CAEE+MSEM) achieves the best PSNR/SSIM on \textsc{MSEC}, while removing either block degrades performance; dropping MSEM hurts more than dropping CAEE, indicating that multi-scale, spectrum-aware injection is the primary driver of fidelity and that CAEE provides a complementary global control signal. 
Figure~\ref{fig:ablation} qualitatively confirms these roles: without CAEE, results show biased global luminance and mild colour casts; without MSEM, local shadows/highlights remain uncorrected despite a valid descriptor; removing both yields the weakest outputs, whereas the complete model preserves structure with balanced exposure. 

\subsection{Computational Efficiency Analysis.} 
As shown in Table~\ref{tab:efficiency}, HICNet achieves real-time inference and significantly outperforms heavy optimization-based baselines such as QuadPrior and ECM. While comparable in size to UEC, our method delivers higher throughput and a superior balance between complexity and performance. This efficiency makes HICNet ideal for resource-constrained and latency-sensitive applications.
\begin{table}[h]
\centering
\vspace{-15pt}
\caption{Computational efficiency on NVIDIA RTX 3090}
\label{tab:efficiency}
\renewcommand{\arraystretch}{0.8}
\setlength{\tabcolsep}{1pt}
\resizebox{0.48\textwidth}{!}{
\begin{tabular}{l c c c c}
\toprule
\textbf{Metric} & QuadPrior~\cite{wang2024zero} & ECM~\cite{eyiokur2022exposure} & UEC~\cite{cui2024unsupervised} & \textbf{HICNet (Ours)} \\
\midrule
\textbf{Speed (FPS)} & 0.23 & 34 & 156 & 244 \\
\textbf{Size (MB)} & 7628.79 & 695.00 & 0.08 & 0.16 \\
\bottomrule
\end{tabular}
}
\vspace{-15pt}
\end{table}
\section{Conclusion}
We presented HICNet, a reference-guided framework for exposure correction that enhances visibility in both overexposed and underexposed regions. By coupling a content-agnostic exposure encoder with a multi-scale exposure-modulation network, the method achieves scene-aware adjustments without requiring paired training data. Experiments on MSEC and LOL show consistent gains in fidelity and perceptual quality, competitive with state-of-the-art supervised approaches. Ablations validate the complementary roles of the encoder and modulation, explaining the robustness of the full model across diverse lighting conditions. Beyond accuracy, HICNet is practical for real-world deployment where ground truth is scarce. Future work will refine the architecture for efficiency, extend to related enhancement tasks, and scale evaluation to more diverse datasets and in-the-wild captures.

%
%
%
\bibliographystyle{IEEEtran}
\bibliography{refs}

\end{document}